# Combined Compromise for Ideal Solution (CoCoFISo): a multi-criteria decision-making based on the CoCoSo method algorithm

**Rôlin Gabriel Rasoanaivo** [0000-0002-2496-2672]

Institut de Recherche en Informatique de Toulouse, IRIT- France

Université Toulouse Capitole, France

Université de Toamasina, Madagascar

Email    Rolin.Rasoanaivo@irit.fr

**Morteza Yazdani** [0000-0001-5526-8950]

Universidad Internacional de Valencia, Spain (*corresponding Autor*)

Email    morteza.yazdani@professor.universidadviu.com

**Pascale Zaraté** [0000-0002-5188-1616]

Institut de Recherche en Informatique de Toulouse, IRIT- France

Université Toulouse Capitole, France

Email    Pascale.Zarate@ut-capitole.fr

**Amirhossein Fateh** [0000-0002-6911-9515]

Universitat Politècnica de València, Spain

Email    afateh@doctor.upv.es

**Abbreviations list**

| | |
|---|---|
| AHP | : Analytical Hierarchy Process |
| BNN | : Bipolar Neutrosophic Numbers |
| BWM | : Best Worst Method |
| CoCoFISo | : Combined Compromise For Ideal Solution |
| CoCoSo | : Combined Compromise Solution |
| CODAS | : COmbinative Distance-based ASsessment |
| COPRAS | : COmplex PRoportional ASsessment |
| CR | : Consistency Ratio |
| CRITIC | : CRiteria Importance Through Intercriteria Correlation |
| CSP | : Cloud Service Providers |
| CSP | : Cloud Service Providers |
| DMU | : Decision-making Units |
| EDAS | : Evaluation based on Distance from Average Solution |
| ELECTRE | : Elimination and Choice Translating Reality |
| L1 | : First year of Bachelor's degree |
| L2 | : Second year of Bachelor's degree |
| LBWA | : Level Based Weight Assessment |
| LSS | : Lean Six Sigma |
| M-AHP | : Modified Analytical Hierarchy Process |
| MABAC | : Multi-Attributive Border Approximation area Comparison |
| MAGDM | : Multiple Attribute Group Decision Making |
| MAHP | : Multiplicative Additive Hierarchical Process |
| MCDM | : Multi-criteria Decision-Making Method |
| MCGDM | : Multiple Criteria Group Decision Making |
| MSME | : Micro, Small, and Medium Enterprises |
| MSW | : Municipal Solid Waste |
| P2TL | : Picture 2-Tuple Linguistic |
| P2TLN | : Picture 2-Tuple Linguistic Numbers |
| P2TLWA | : Pythagorean 2-Tuple Linguistic Weighted Average |
| P2TLWG | : Pythagorean 2-Tuple Linguistic Weighted Geometric |
| PDHLTS | : Probabilistic Double Hierarchy Linguistic Term Set |
| PEE | : Productive Economic Endeavours |
| PMVNN | : Probability multi-valued neutrosophic numbers |
| PMVNS | : Probability Multi-Valued Neutrosophic Set |
| PROMETHEE | : Preference Ranking Organisation Method for Enrichment Evaluations |
| RDF | : Refuse Derived Fuel |
| SA | : Sentiment Analysis |
| SAHP | : Simple Additive Weighting Hierarchical Process |
| SPBRT | : Spent Power Battery Recycling Technology |
| SVN2TLFWA | : Single-Valued Neutrosophic Two-Tuple Linguistic Frank weighted averaging |
| SVN2TLFWG | : Single-Valued Neutrosophic Two-Tuple Linguistic Frank weighted geometric |
| SVN2TLS | : Single-Valued Neutrosophic Two-Tuple Linguistic Set |
| T-SFFS | : T-spherical fuzzy Frank Softmax |
| TOPSIS | : Technique for Order Preference by Similarity to the Ideal Solution |
| VIKOR | : VlseKriterijuska Optimizacija I Komoromisno Resenje |
| WSM | : Weighted Sum Method |

**Abstract**

Each decision-making tool should be tested and validated in real case studies to be practical and fit to global problems. The application of multi-criteria decision-making methods (MCDM) is currently a trend to rank alternatives. In the literature, there are several multi-criteria decision-making methods according to their classification. During our experimentation on the Combined Compromise Solution (CoCoSo) method, we encountered its limits for real cases. The authors examined the applicability of the CoCoFISo method (improved version of combined compromise solution), by a real case study in a university campus and compared the obtained results to other MCDMs such as Preference Ranking Organisation Method for Enrichment Evaluations (PROMETHEE), Weighted Sum Method (WSM) and Technique for Order Preference by Similarity to the Ideal Solution (TOPSIS). Our research finding indicates that CoCoSo is an applied method that has been developed to solve complex multi variable assessment problems, while CoCoFISo can improve the shortages observed in CoCoSo and deliver stable outcomes compared to other developed tools.

The findings imply that application of CoCoFISo is suggested to decision makers, experts and researchers while they are facing practical challenges and sensitive questions regarding the utilization of a reliable decision-making method. Unlike many prior studies, the current version of CoCoSo is unique, original and is presented for the first time. Its performance was approved using several strategies and examinations.

**Keywords: MCDM, CoCoSo method, CoCoFISo method**

## 1. Introduction

Multiple criteria decision making has become a wider area of research for different groups of researchers in various disciplines from supply chain and manufacturing to the business and management, energy, or construction topics Kumar et al. (2017); Fallahpour et al. (2021); Torkayesh et al. (2022). During the last thirty years, a range of decision-making tools were developed to find the optimal solution for multi criteria decision making problems. From the creation of Analytical Hierarchy Process (AHP) by Saaty and ELECTRE or PROMETHEE by Roy to the formulation of novel methods like Best Worst Method (BWM), investigators in this field have tried to improve and extend the structure and anatomy of MCDM methods Govindan et Jepsen, (2016); Govindan et al. (2017); Khan et al.( 2023). As it is known, some methods are utilized to find the weights of decision criteria, however there are other methods that just will be utilized for alternative comparison and ranking. From this category we can name TOPSIS, VIKOR, COPRAS and a recent method Combined Compromise Solution (CoCoSo). Rather than its applicability and unique algorithm, it can meet decision makers requirements and deliver a dynamic environment to the decision-making problem Yazdani et al. (2019a); Kumar et al. (2022).

The fast-growing application of CoCoSo is visible through the MCDM literature. Due to its understandable algorithm and less complexity in implementation, experts in different sectors try to conduct CoCoSo in their research. For instance, Ecer et Pamucar, (2020) proposed a fuzzy integration of BWM and CoCoSo with Bonferroni concept to validate the interaction between decision variables and reduce the influence of extreme/awkward data on the values of the criterion functions and checking the robustness of the results through a variation of the CoCoSo Bonferroni parameters. Yazdani et al. (2019b) developed a grey interval extension of CoCoSo to evaluate the performance of construction suppliers. Torkaysh et al. (2021) conducted a combined framework of BWM-LBWA-CoCoSo to European healthcare system evaluation. CoCoSo has received notable attention among the decision making community. It obtained considerable citations among the research community and it is implemented in many applications such as transport and logistics to circular economy and finance etc. Khan et Haleem, (2021); Pen and Huang, (2020); Ulutaş et al. (2020). The original version of the CoCoSo method was published in management science journal, 2019.

The main idea of this work is not to add complexity to the first version of the method, but it is to correct one step of the method that was correct for some specific cases. About complexity, we make a distinction between four forms of complexity that also help shed light on different dimensions of systems thinking: dynamic, architectural, relational and generative complexity. This work has no link with these four kinds of complexity.  In the next section, we provide a wider review of the improved version of the MCDM and comparison of them. Then a problem statement and research contribution will be explained. Section 3 and 4 provides the improvement of the CoCoSo method through several examples and real case problems. In section 5, we perform several tests like sensitivity analysis of weights and comparison with other MCDM tools. At the end, the conclusion and discussion are presented.

## 2. Literature review and Problem statement

This section provides a survey of the history of the MCDM method improvement (extension), and comparing various models based on their classical or improved version. At the end we develop the problem statement and research contributions.

### *2.1. Highlighted MCDM improvement (extension) summary*

Along with the development of the recent MCDM methods like Best Worst Method (BWM), EDAS, CODAS, several studies dedicated an improvement or extension of the MCDM tools to check the performance of them. For example, Nefeslioglu et al. (2013) introduced a modified Analytical Hierarchy Process (M-AHP) with a corresponding computer code. The conventional AHP has been widely used in various application areas, but its main advantage and disadvantage lie in the uncertainties stemming from expert judgments. The study proposes integrating fuzzy logic into AHP (Fuzzy-AHP) to handle uncertainties. The M-AHP in this study focuses on expert-driven comparisons of factors and decision points, terminating the expert's role after determining total scores and instant assignments for relevant cases. Notably, the M-AHP methodology ensures that the consistency ratio (CR) never exceeds 0.10, enhancing its practicality. A case study evaluating snow avalanche susceptibility demonstrates the application of M-AHP, but the technique can be applied to various natural hazards. The study emphasizes evaluating all relevant terrain mapping units for a comprehensive hazard map. Finally, a computer code for M-AHP is presented, serving as a valuable tool for researchers dealing with natural hazards, natural resources, or nature preservation and employing the M-AHP technique for decision support. Tung and Tang (1992) conducted an analysis on 147 reciprocal matrices (clusters) to observe the performance of eigenvector approaches in the Analytic Hierarchy Process (AHP). The observations revealed several points as follow: For pre-arranged order of objects, both the Simple Additive Weighting Hierarchical Process (SAHP) and Multiplicative Additive Hierarchical Process (MAHP) approaches could successfully yield the correct order of alternatives in the overall ranking, with some inconsistencies and rank reversals in individual clusters. The SAHP showed fewer rank reversals (6 sets) compared to the MAHP (18 sets) between the right and left eigenvector approaches. Under less-obvious order of objects, the SAHP yielded more consistent overall ranking results (6 out of 21 models) using both eigenvector approaches compared to the MAHP (9 out of 21 models). Full hierarchical analysis exhibited a higher occurrence of rank reversals compared to clustering, especially with a greater number of levels in the hierarchical structure. Ultimately, the study concluded that the MAHP is not superior to the SAHP in solving AHP problems, particularly based on the observed rank reversals in the MAHP.

Another work introduced an Adaptive Analytic Hierarchy Process (AHP) Approach (A3) aiming to enhance the traditional AHP method for solving Multiple Criteria Decision Making (MCDM) problems. The A3 approach brings improvements in three main areas: cost-effectiveness, by eliminating the reassessment process and saving significant man-hours and associated costs; timeliness, by significantly reducing data-collection time, making it feasible to address emergent MCDM problems promptly; and improved decision quality, ensuring a more accurate reflection of decision-makers' beliefs regarding the relative importance relationships of criteria. The case study demonstrated substantial savings in man-hours (42.6%) and data-collection time (68%) using the A3 approach. Future research focuses on addressing limitations and developing an automated system for A3 to further expedite data collection Lin et al. (2008). Yadav G. et al. (2018) focused on guiding the prioritization and ranking of solutions to address barriers and facilitate the adoption of Lean Six Sigma (LSS). The approach involves identifying LSS barriers and solutions through a literature review and expert input. The study proposed a hybrid framework utilizing fuzzy set theory, fuzzy Analytical Hierarchy Process (AHP), and modified TOPSIS (Technique for Order Preference by Similarity to Ideal Solution). Sensitivity analysis is employed to validate the framework's robustness. The findings highlight key LSS barriers, solutions, and a robust framework to guide LSS adoption effectively.

Tansel Y. et al. (2022) introduced a financial performance measurement model tailored for the wholesale and retail trade sector, utilizing the AHP-modified VIKOR method. This model proved valuable for performance analysis when relevant financial performance measurement criteria are available. By combining the strengths of AHP and modified VIKOR methods, the model addressed their respective disadvantages, offering a comprehensive framework for financial performance evaluation. The modified VIKOR method, without requiring a hierarchical decision-making structure, evaluated a company's performance based on directly assigned criteria weights, a crucial factor in company rankings. The study proposed a hierarchical structure using the AHP model for criteria weighting, allowing independent evaluation of key criteria like liquidity, financial statement, and profit ratios. This structured approach enhanced the financial performance measurement process. Compared to other Multi-Criteria Decision Making (MCDM) methods, this model considers multiple-independent financial ratio data and their weight impacts, providing insights into consensus ranking results from various financial perspectives. The study applied this integrated model to rigorously measure financial performance in the wholesale and retail sector, aiming to offer valuable insights to company managers and investors. The AHP-modified VIKOR integrated model offered flexibility, enabling periodic monitoring based on data structures and facilitating competitive financial performance measurement and strategic improvement.

Zhang A. et al. (2021) explored various weighting methods for evaluating indicators from different perspectives and employed game theory to find an optimal combination of index weights, addressing issues related to subjectivity and imbalanced weight ratios. Additionally, the introduction of prospect theory accounted for decision makers' psychological perceptions, enhancing the traditional VIKOR method and achieving a balance between subjective and objective factors. Example results demonstrated the effectiveness and reasonableness of the proposed evaluation model, which outperforms existing methods and offers a scientific basis for assessing black-start schemes. However, it's worth noting that this paper only used the analytic hierarchy process for subjective weighting, which may not be suitable for black-start schemes with numerous evaluation indicators. Gou et al. (2020) introduced a new and comprehensive concept termed "probabilistic double hierarchy linguistic term set" (PDHLTS). The study concentrates on defining rational operations and a distance measure for PDHLTSs, presenting an adjustment method to ensure identical probability distribution for two PDHLTSs. Furthermore, the extended probabilistic double hierarchy linguistic VIKOR method, is developed by refining the traditional VIKOR method. The effectiveness and practicality of this method are showcased through its application to resolve a real-world multiple criteria decision-making problem related to smart healthcare. The paper concluded with comparative analyses and discussions on potential directions for future research.

Zhang C, et al. (2020) proposed a new method for ranking products based on prospect theory, sentiment analysis (SA), and an enhanced version of VIKOR. The method involves three main stages: screening alternative products, calculating sentiment scores, and ranking the products. A case study on automobile selection validates the effectiveness of this approach in aiding consumers to choose products that align with their preferences. Aghajani Mir et al. (2016) extensively examined municipal waste management methods, particularly focusing on recycling, composting, and RDF (Refuse Derived Fuel) conversion, evaluating their viability and environmental impacts for future implementation. The research offered recommendations to waste managers for enhancing waste management systems through sustainable approaches. The evaluation involved comparing and ranking various MSW (Municipal Solid Waste) management scenarios based on economic and environmental criteria. The methodology utilized MCDM (Multi-Criteria Decision Making) techniques, including an improved version of TOPSIS (Technique for Order of Preference by Similarity to Ideal Solution). The ranking process also incorporated sensitivity analysis and a combination of TOPSIS and VIKOR methods. The results identified the most suitable and inappropriate scenarios for municipal waste management, highlighting the current inadequacies in the existing conditions of municipal SWM (Solid Waste Management). Wang and Wang (2014) addressed the evaluation of competitiveness in the Chinese high-tech industry, recognizing it as a crucial aspect of national strength. It highlighted the challenges of evaluating high-tech industries, emphasizing the common issue of conflict and correlation among evaluation indices. The proposed solution was an improved Technique for Order of Preference by Similarity to Ideal Solution (TOPSIS) method based on the Mahalanobis distance concept, aiming to mitigate the impact of indicator correlation. The improved TOPSIS method was validated and shown to offer a more accurate tool for Multi-Criteria Decision Making (MCDM) problems, providing a scientific basis for decision-making and aligning with practical evidence, especially in the context of Chinese high-tech industrial competitiveness.

Rao and Gao (2022) addressed the pressing issue of promoting urban-rural integration development in the context of new urbanization and rural revitalization strategy. Using index data from 15 cities in Hubei Province, the paper proposed a novel evaluation mechanism to assess the urban-rural integration development level. The main contributions included the establishment of a comprehensive index system, determination of index weights using a combination weighting method, and the introduction of an improved TOPSIS method based on relative entropy and grey relational degree for evaluation. Empirical analysis on Hubei Province's cities in 2020 was conducted, leading to insights into urban-rural integration development status. The study suggested effective policy implications to enhance urban-rural integration in Hubei Province. While the paper identified areas for improvement, such as incorporating data from more years for trend analysis and refining evaluation regions to a finer level, it successfully laid a foundation for further research and analysis in the domain of urban-rural integration development.

Chena W. et al. (2018) developed a comprehensive index system to evaluate the transformation and upgrading of resource-based cities in Shaanxi province from 2001 to 2015, considering economic adjustment, social progress, and environmental improvement. An enhanced evaluation method called TOPSIS, along with sequence weighting, was used for this assessment. The study's conclusion provided valuable reference points for the government to establish a comprehensive index system and assess the transformation and upgrading of resource-based cities. It emphasized the need for continued efforts to transition from resource-dependent development to sustainable development, integrating resource conservation and environmental protection throughout the economic and social processes. Wu et al. (2021) worked on the safety evaluation of urban rail stations and systematically developed an evaluation index system for this purpose. To improve the evaluation of urban rail transit station operation safety, the paper enhances and optimizes the entropy weight method. It emphasizes the utilization of original data, respecting objective realities, and proposing improvement steps to ensure that the obtained index weights align better with the actual situation. The traditional TOPSIS method is also improved by introducing a virtual worst solution to provide more accurate ranking results, effectively reflecting the strengths and weaknesses of each evaluated scheme. This multi-index and multi-objective analysis combines the advantages of the TOPSIS method, and the improved entropy weight method replaces subjective weighting methods, enhancing the rationality of index weighting and the objectivity of the TOPSIS method. The main research outcomes of this paper included the development of an evaluation index system for urban rail station operation safety, with clear definitions and detailed calculation methods. The proposed evaluation method, based on the improved TOPSIS and entropy weight methods, is efficient, scientifically sound, and practical. Case analysis results demonstrated that this evaluation system can systematically and comprehensively assess the safety of urban rail stations, providing a solid foundation for urban rail transit planning and management.

Wu J. et al. (2011) addressed limitations in evaluating and ranking decision-making units (DMUs) using average cross-efficiency scores. Instead of relying on averages, the study utilized the TOPSIS technique to rank cross-efficiency, aiming to overcome these limitations. The proposed approach is demonstrated through a numerical example to showcase its effectiveness. The resulting rankings utilized original data from the cross-efficiency matrix, providing a quantitative reflection of the strengths and weaknesses of various evaluation programs. The numerical example was primarily for illustration, and the authors suggested further exploration and application of this approach in real-world scenarios for a more comprehensive understanding of its potential benefits. Irvanizam et al. (2021) introduced an improved version of the Evaluation Based on Distance from Average Solution (EDAS) method for Multiple Criteria Group Decision Making (MCGDM) problems represented by Bipolar Neutrosophic Numbers (BNNs). BNNs were deemed effective for expressing cases involving vagueness, inconsistency, incompleteness, and uncertainty often encountered in decision-making scenarios. The enhanced EDAS calculated the average alternative by aggregating BNNs for each criterion using specific operators. It then computed positive and negative distances between each evaluated alternative and the average solution. The method adjusted parameter values based on decision-makers' preferences and ranked the alternatives accordingly. A case study evaluating the best MSME for distributing the PEE program demonstrated the stability and effectiveness of the proposed EDAS method in handling MCGDM problems with BNNs. The method required less preprocessing and computation, making it suitable for scenarios with conflicting criteria and alternatives. However, a challenge for future work lies in establishing strategies to handle unknown weight matrices to achieve more realistic decision results. Furthermore, there's potential to extend the EDAS method to other neutrosophic and fuzzy set variations. Peng D. et al. (2022) proposed an enhanced evaluation method, EDAS, specifically for interval-valued intuitionistic trapezoidal fuzzy numbers. The key contributions were twofold: firstly, considering decision makers' expectation levels regarding emergency plans to better handle dynamic plan development; and secondly, utilizing a new distance measure between these fuzzy numbers to calculate decision makers' expected levels, resulting in emergency plans that align more closely with reality. However, the paper identified limitations such as not addressing the consistency of decision makers' information and suggesting potential areas of future research. These included exploring information consistency among multiple decision makers, optimizing the consistency adjustment of evaluation information, extending EDAS to continuous fuzzy information and multi-source data, and applying the method to various domains like investment decision-making and group decision-making in emergency scenarios. Zhang S. et al. (2021) introduced a novel method, combining CODAS (Complex Proportional Assessment) with Picture 2-Tuple Linguistic Numbers (P2TLNs), to address Multiple Attribute Group Decision Making (MAGDM) challenges. The methodology involved reviewing fundamental information about P2TLSs, integrating them using P2TLWA and P2TLWG operators, determining attribute weights using the CRITIC method, and applying CODAS in a picture 2-tuple linguistic environment. An application related to selecting the optimal green supplier confirms the method's validity. Comparative analyses with other methods highlighted the advantages of this approach. The contributions of this research include modifying the CODAS method using P2TLNs and designing the P2TL-CODAS method for MAGDM with P2TLNs. The CRITIC method is utilized for attribute weight determination, and a case study demonstrated the effectiveness of the developed method. In future research, these proposed methods and algorithms can be extended to address various real decision-making problems involving fuzzy and uncertain information. Lei Xu et al. (2021) explored Multiple Attribute Group Decision Making (MAGDM) within a Single-Valued Neutrosophic Two-Tuple Linguistic Set (SVN2TLS) environment. The study extended

Frank triangular norms to SVN2TLS, introducing new operational rules, and investigating related properties. The enhanced Frank operation allowed for flexible parameter selection based on actual situations and decision makers' preferences, improving the assembly process. Additionally, two new aggregation operators (SVN2TLFWA and SVN2TLFWG) were proposed, and their desirable properties were discussed. The study also integrated the improved multi-attributive border approximation area comparison (MABAC) method with SVN2TLNs information to propose two approaches for solving MAGDM problems, with detailed computing steps provided. An iterative algorithm based on the BWM method was developed for determining unknown expert weights, and risk attitude parameters were introduced to make the method adaptable to decision makers' risk attitudes during assessment. The paper concluded with examples demonstrating the proposed method's advantages and outlines future research directions, including extending the proposed operators to large group decision models and algorithms like consensus and clustering models. Liu P. et al. (2022) enhanced the multi-attributive border approximation area comparison (MABAC) method within the Probability Multi-Valued Neutrosophic Set (PMVNS) framework, introducing a three-phase multi-attribute group decision-making (MAGDM) method. The study began by reviewing PMVNS concepts, traditional MABAC method, and regret theory (RT). A similarity measure for PMVNSs was defined to calculate the important degree of decision makers (DMs). The likelihood of preference relations using probability multi-valued neutrosophic numbers (PMVNNs) was introduced to replace distance deviation in the traditional MABAC method. The proposed MAGDM method expresses alternative performance using PMVNNs, integrating the likelihood-based MABAC method and RT that considered DMs' behavioral psychology. A case study was conducted to illustrate the feasibility and applicability of the proposed approach.

Han Lai et al. (2020) proposed an enhanced CoCoSo (Combined compromised solution) method for selecting Cloud Service Providers (CSPs). The original CoCoSo method had a flaw in giving equal importance to three subordinate compromise scores. To address this, the study introduced "Linear Sum Normalization" to balance the aggregated results. Additionally, a nonlinear programming model with variance maximization was introduced to enhance discrimination in the evaluation results and aid decision-makers in selecting the optimal CSP. The proposed method was validated through a trust evaluation example using synthetic and real cloud data. The results demonstrated stability, consistency with existing decision-making methods, and effectiveness in identifying subtle differences among alternatives. Future work will focus on extending the algorithm using Z-numbers and D-numbers to handle information reliability problems in the decision-making process. The aim was to enhance evidential reasoning and accuracy in CSP selection, with potential applications in renewable energy investment and green economy development, considering sustainability and multiple conflicting criteria. Wang H. et al. (2023) presented a novel CoCoSo (Combined Compromise Solution) method that leveraged Frank operational laws and the softmax function to address multiple attribute group decision-making problems using T-spherical fuzzy sets. The study extended Frank operations to a T-spherical fuzzy environment and introduced aggregation operators like T-spherical fuzzy Frank softmax (T-SFFS) average and geometric operators, along with their weighted forms. The properties and specific cases of these operators were discussed, and their monotonicity is analyzed to indicate decision-makers' decision attitudes and risk preferences. The article also proposed a CoCoSo method based on Hamming distance measure, considering decision-makers' attitudes and attribute priority. A multiple attribute group decision-making framework employing two independent and parallel T-spherical fuzzy information processing is designed. The proposed method's practicality is demonstrated through a case study on spent power battery recycling technology (SPBRT) selection. Sensitivity and comparative analyses confirm the reliability, effectiveness, and superiority of the proposed method.

### 2.2. Research gap and contribution

Development of the MCDM methods is growing and this has been reflected widely in the literature and many researchers indicated their point of view about improved versions of MCDM methods according to the above references. This research conducts a problematic issue in the CoCoSo method when in special cases the algorithm receives an error and it is blocked. Based on our observation two conditions were assumed. On one side when a criterion has the same value for all the alternatives, we are not able to compute the normalization process. On the other hand, we observed that when an alternative has the worst item in all the criteria, it will not be possible to compute the aggregation step in CoCoSo and the algorithm. It is understandable both situations and the received errors, for the first error that criterion can be eliminated and in the second one that alternative will be the worst item and we can ignore it from the alternative list. However, those errors are observable if we have limited alternatives and in case there are more than 30 or 40 alternatives (even more), finding those alternatives and criteria will be complicated. This is the research gap we encountered, and in this research, we provide some strategies to improve and refuse those errors. In addition, there is no study that proposes a new algorithm for CoCoSo and in this aspect, to the best of our knowledge and survey, for the first time such a method is extended with extensive analysis. Therefore, the contribution of this study is to propose some reconfigurations in the CoCoSo method to reduce probable errors and then comparing the improved version with other tools like PROMETHEE, WSM and TOPSIS. In the next section, we describe completely the observed errors with a real case example for student resident allocation and propose our new version of CoCoSo called CoCoFISo. Then sort of sensitivity analysis tests and comparison will be accomplished. At the end we present a discussion and conclusion.

## 3. Improvement on CoCoSo algorithm

### 3.1. Classic CoCoSo model

CoCoSo starts to find the most appropriate alternative by combining ideas of compromised solutions like mean evaluation weighting and power weight aggregation. The step-by-step solution of CoCoSo is interpreted here:

1. First step in this method is to form the decision problem including determination of alternatives, criteria, the weights of each criterion, their optimization direction. Then, we should form an initial decision matrix. This action should be performed by accessing a data set or decision-making preference (utilizing linguistic values).

2. Second step is to normalize the matrix; this is done through two strategies for the benefit and non-benefit category of criteria like price of a product or energy consumption etc. This is handled through formulas 2 and 3 respectively.

3. CoCoSo determines two strategies to aggregate weights of the criteria in the decision process, including formulas 4 and 5. One is to sum the multiplication of the normalized matrix by weights values (S) and second one is to sum the power weight of the normalized matrix (P).

4. By this step we must integrate S and P values. This step is managed through the formulas 6, 7 and 8. Formula 6 is the arithmetic mean of sums of WSM and WPM scores, while Equation (7) expresses a sum of relative scores of WSM and WPM compared to the best. Equation (8) releases the balanced compromise of WSM and WPM models scores.

5. The final ranking of the alternatives is determined through k values in equation 9.

The applicability of CoCoSo is compared to other similar tools or sensitivity analysis tests and this aids decision experts to validate the results. Here is the mathematical formulation for CoCoSo implementation.

a) Determining the initial decision-making matrix as shown below:

$$x_{ij} = \begin{bmatrix} x_{11} & x_{12} & \cdots & x_{1n} \\ x_{21} & x_{22} & \cdots & x_{2n} \\ \cdots & \cdots & \cdots & \cdots \\ x_{m1} & x_{m2} & \cdots & x_{mn} \end{bmatrix}; i=1,2,\ldots,m; \; j=1,2,\ldots,n. \tag{1}$$

b) The normalization of criteria values is accomplished based on the compromise normalization equation (see Zeleny (1973)).

$$r_{ij} = \frac{x_{ij} - \min_i x_{ij}}{\max_i x_{ij} - \min_i x_{ij}}; for\ benefit\ criterion; \tag{2}$$

$$r_{ij} = \frac{\max_i x_{ij} - x_{ij}}{\max_i x_{ij} - \min_i x_{ij}}; for\ cost\ criterion \tag{3}$$

c) Obtain the total of the weighted comparability sequence and the whole of the power weight of comparability sequences for each alternative sum of the weighted comparability sequence and an amount of the power weight of comparability sequences for each alternative as $S_i$ and $P_i$, respectively:

$$S_i = \sum_{j=1}^{n}(w_j r_{ij}). \tag{4}$$

This $S_i$ value achieved based on the grey relational generation approach.

$$P_i = \sum_{j=1}^{n} (r_{ij})^{w_j} \qquad (5)$$

This *P$_i$* value is also achieved according to the WASPAS multiplicative attitude.

d) Compute relative weights of the alternatives using the following aggregation strategies. In this step, three appraisal score strategies to generate relative weights of other options, which are derived by formulas (6), (7), and (8), are proposed.

A) $k_{ia} = \dfrac{P_i + S_i}{\sum_{i=1}^{m}(P_i + S_i)}$ \hfill (6)

B) $k_{ib} = \dfrac{S_i}{\min_i S_i} + \dfrac{P_i}{\min_i P_i}$ \hfill (7)

C) $k_{ic} = \dfrac{\lambda(S_i) + (1-\lambda)(P_i)}{(\lambda \max_i S_i + (1-\lambda) \max_i P_i)}; 0 \leq \lambda \leq 1.$ \hfill (8)

In equation (8) $\lambda$ (usually $\lambda = 0.5$) is chosen by decision-makers. However, the flexibility and stability of the proposed *CoCoSo* can rely on other values.

e) The final ranking of the alternatives is determined based on *k$_i$*, values (as more significant as better):

$$k_i = (k_{ia} k_{ib} k_{ic})^{\frac{1}{3}} + \frac{1}{3}(k_{ia} + k_{ib} + k_{ic}) \qquad (9)$$

### *3.2. Real case example and Discussion*

Since the development of the CoCoSo method, hundreds of studies incorporated various versions of that and applied it in different case studies, From a classical or original CoCoSo to grey or interval models and some fuzzy extensions. Researchers focused on the utility of the CoCoSo in uncertain conditions and took the advantage of its user-friendly and easy-going algorithm. We have reviewed all the resources that used CoCoSo in their application, to the best of our knowledge, none of them approached a modified or improved version of CoCoSo. Although this is a comprehensive full ranking method and the structure has been tested in many ways, we noticed several errors while applying the CoCoSo method on special occasions. We will present and discuss this situation below.

During research on using multi-criteria decision-making methods to select students and to allocate them university accommodation, we used several methods including CoCoSo Rasoanaivo, (2023). The objective is to sort students based on their qualifications. Five criteria to be able to select them according to the availability of accommodation. Clearly, it is impossible to accommodate all students in university residences due to the lack of student accommodation. For this objective, there are social criteria chosen by universities to enable student assessment. We adopted the case of the University of Toamasina in Madagascar which uses five (5) social criteria when selecting students including:

- PC: physical capacity of the student;
- OP: orphanage situation of the student's parent;
- PW: parent's profession condition;
- DC: number of dependent children of the parent;
- DR: distance of student's main residence from the university.

Table 1 below describes these criteria and their value depending on the case.

**Table 1** – the evaluation criteria for university students

| Social criteria | Value |
|---|---|
| PC | Normal = 5 ; Disability = 10 |
| OP | None =5; Father or Mother = 10 ; Father and Mother = 15 |
| PW | University = 10 ; Other = 5 |
| DC | By number |
| DR | By mileage |

The priority level of these criteria is as follows. Initially, the university prioritizes students who have a physically handicapped ability at the first level. Then, the orphan parent criterion and the student's homeland distance from the university are secondly prioritized. Finally, the number of dependent children of the parent and the parent's place of work have the same priority. Therefore, the weights of the social criteria are presented in the following:

$W_{pc}$: 0.45     $W_{OP}$: 0.18     $W_{PW}$: 0.1     $W_{DC}$: 0.1     $W_{DR}$: 0.18

Note that these criteria weights were obtained according to the own choice of the decision-maker (Directorate of University Works) within the University of Toamasina by applying the AHP method Rasoanaivo, (2023).

The social situation of students is treated by training and level of study. Thus, the ordering must be carried out by training and level of study as well.

Here we propose an example of students from the Faculty of Letters who follow the training in French language studies in the first (L1) and second (L2) year of the Bachelor's degree. The following table presents the social situation in value of these students (Table 2).

**Table 2** – first and second group of students in French language studies

| L1 Students | | | | | | L2 Students | | | | | |
|---|---|---|---|---|---|---|---|---|---|---|---|
| Student | PC | DR | DC | PW | OP | Student | PC | DR | DC | PW | OP |
| L101 | 5 | 100 | 3 | 5 | 5 | L201 | 5 | 757 | 3 | 5 | 5 |
| L102 | 5 | 100 | 5 | 5 | 10 | L202 | 5 | 102 | 4 | 5 | 10 |
| L103 | 5 | 100 | 6 | 5 | 5 | L203 | 5 | 100 | 4 | 5 | 5 |
| L104 | 5 | 102 | 2 | 5 | 10 | L204 | 5 | 100 | 3 | 5 | 10 |
| L105 | 5 | 100 | 3 | 5 | 5 | L205 | 5 | 100 | 2 | 5 | 10 |
| L106 | 10 | 100 | 5 | 5 | 10 | L206 | 5 | 100 | 2 | 5 | 10 |
| L107 | 5 | 100 | 4 | 5 | 10 | L207 | 5 | 923 | 3 | 5 | 10 |
| L108 | 5 | 100 | 2 | 5 | 5 | L208 | 5 | 1436 | 2 | 10 | 5 |
| L109 | 5 | 100 | 3 | 5 | 5 | L209 | 5 | 1465 | 3 | 5 | 5 |
| L110 | 5 | 923 | 1 | 5 | 15 | L210 | 10 | 100 | 3 | 5 | 5 |
| L111 | 5 | 100 | 3 | 5 | 5 | L211 | 5 | 757 | 3 | 5 | 5 |
| L112 | 10 | 100 | 4 | 5 | 10 | L212 | 5 | 399 | 2 | 5 | 10 |
| L113 | 5 | 100 | 2 | 5 | 5 | L213 | 5 | 100 | 5 | 5 | 10 |
| L114 | 5 | 100 | 4 | 5 | 5 | L214 | 5 | 102 | 3 | 5 | 5 |
| L115 | 5 | 100 | 1 | 5 | 10 | L215 | 5 | 1275 | 3 | 5 | 5 |
| L116 | 5 | 350 | 2 | 5 | 15 | L216 | 5 | 350 | 3 | 5 | 5 |
| L117 | 5 | 100 | 6 | 5 | 5 | L217 | 5 | 350 | 2 | 5 | 10 |
| L118 | 5 | 100 | 5 | 5 | 5 | L218 | 10 | 100 | 2 | 5 | 10 |
| L119 | 10 | 100 | 2 | 5 | 5 | L219 | 5 | 102 | 3 | 10 | 10 |
| L120 | 5 | 102 | 1 | 5 | 5 | L220 | 5 | 100 | 2 | 5 | 5 |
| L121 | 5 | 100 | 4 | 5 | 10 | L221 | 5 | 100 | 1 | 5 | 5 |
| L122 | 5 | 100 | 5 | 5 | 10 | L222 | 5 | 100 | 3 | 5 | 5 |
| L123 | 5 | 399 | 1 | 5 | 5 | L223 | 5 | 1275 | 4 | 5 | 5 |
| L124 | 5 | 100 | 3 | 5 | 15 | L224 | 5 | 609 | 3 | 5 | 5 |
| L125 | 5 | 923 | 4 | 5 | 10 | L225 | 5 | 100 | 2 | 5 | 5 |
| L126 | 5 | 399 | 2 | 5 | 15 | L226 | 5 | 920 | 3 | 10 | 5 |
| L127 | 5 | 100 | 2 | 5 | 5 | | | | | | |

We will now proceed to apply the CoCoSo method algorithm step by step to detect errors.

The first step of the CoCoSo method algorithm is to determine the decision matrix (a). Next step is to normalize the matrix which consists of standardizing this decision matrix (b). Table 3 gives us the result of this normalization of the decision matrix for students in the first year and second year (**L1 and L2**).

**Table 3** – The normalized decision matrix

| L1 Students | | | | | | L2 Students | | | | | |
|---|---|---|---|---|---|---|---|---|---|---|---|
| Student | PC | DR | DC | PW | OP | Student | PC | DR | DC | PW | OP |
| L101 | 0 | 0 | 0,4 | #DIV/0! | 0 | L201 | 0 | 0,48 | 0,5 | 0 | 0 |
| L102 | 0 | 0 | 0,8 | #DIV/0! | 0,5 | L202 | 0 | 0 | 0,75 | 0 | 1 |
| L103 | 0 | 0 | 1 | #DIV/0! | 0 | L203 | 0 | 0 | 0,75 | 0 | 0 |
| L104 | 0 | 0 | 0,2 | #DIV/0! | 0,5 | L204 | 0 | 0 | 0,5 | 0 | 1 |
| L105 | 0 | 0 | 0,4 | #DIV/0! | 0 | L205 | 0 | 0 | 0,25 | 0 | 1 |
| L106 | 1 | 0 | 0,8 | #DIV/0! | 0,5 | L206 | 0 | 0 | 0,25 | 0 | 1 |
| L107 | 0 | 0 | 0,6 | #DIV/0! | 0,5 | L207 | 0 | 0,6 | 0,5 | 0 | 1 |
| L108 | 0 | 0 | 0,2 | #DIV/0! | 0 | L208 | 0 | 0,98 | 0,25 | 1 | 0 |
| L109 | 0 | 0 | 0,4 | #DIV/0! | 0 | L209 | 0 | 1 | 0,5 | 0 | 0 |
| L110 | 0 | 1 | 0 | #DIV/0! | 1 | L210 | 1 | 0 | 0,5 | 0 | 0 |
| L111 | 0 | 0 | 0,4 | #DIV/0! | 0 | L211 | 0 | 0,48 | 0,5 | 0 | 0 |
| L112 | 1 | 0 | 0,6 | #DIV/0! | 0,5 | L212 | 0 | 0,22 | 0,25 | 0 | 1 |
| L113 | 0 | 0 | 0,2 | #DIV/0! | 0 | L213 | 0 | 0 | 1 | 0 | 1 |
| L114 | 0 | 0 | 0,6 | #DIV/0! | 0 | L214 | 0 | 0 | 0,5 | 0 | 0 |
| L115 | 0 | 0 | 0 | #DIV/0! | 0,5 | L215 | 0 | 0,86 | 0,5 | 0 | 0 |
| L116 | 0 | 0,3 | 0,2 | #DIV/0! | 1 | L216 | 0 | 0,18 | 0,5 | 0 | 0 |
| L117 | 0 | 0 | 1 | #DIV/0! | 0 | L217 | 0 | 0,18 | 0,25 | 0 | 1 |
| L118 | 0 | 0 | 0,8 | #DIV/0! | 0 | L218 | 1 | 0 | 0,25 | 0 | 1 |
| L119 | 1 | 0 | 0,2 | #DIV/0! | 0 | L219 | 0 | 0 | 0,5 | 1 | 1 |
| L120 | 0 | 0 | 0 | #DIV/0! | 0 | L220 | 0 | 0 | 0,25 | 0 | 0 |
| L121 | 0 | 0 | 0,6 | #DIV/0! | 0,5 | L221 | 0 | 0 | 0 | 0 | 0 |
| L122 | 0 | 0 | 0,8 | #DIV/0! | 0,5 | L222 | 0 | 0 | 0,5 | 0 | 0 |
| L123 | 0 | 0,36 | 0 | #DIV/0! | 0 | L223 | 0 | 0,86 | 0,75 | 0 | 0 |
| L124 | 0 | 0 | 0,4 | #DIV/0! | 1 | L224 | 0 | 0,37 | 0,5 | 0 | 0 |
| L125 | 0 | 1 | 0,6 | #DIV/0! | 0,5 | L225 | 0 | 0 | 0,25 | 0 | 0 |
| L126 | 0 | 0,36 | 0,2 | #DIV/0! | 1 | L226 | 0 | 0,6 | 0,5 | 0 | 0 |
| L127 | 0 | 0 | 0,2 | #DIV/0! | 0 | | | | | | |

While using excel software and comparing the normalized values, we noticed that for L1 students the normalization of the values of the **PW** criteria causes some errors. This means that for L1 students, we are not able to proceed the following steps of applying the CoCoSo method. This happened because the **PW** criterion has the same value for all students. Although in real world cases when a criterion has the same value for all the alternatives, it can be eliminated, however, this blocks us to go forward and achieve the results. Thus, we can draw a first error, (#DIV/0!) meaning that the calculation is not possible for all the PW column due to the division by zero, on the CoCoSo method algorithm as shown in Table 3. However, if we take this same case, it will still be possible to determine the normalized matrix by applying other multi-criteria decision support methods such as AHP, WSM, PROMETHEE Rasoanaivo et Zaraté, (2022). On the other hand, for L2 students, the normalization of values on all criteria went well. Therefore, we continue the computation for the case of L2 students.

The step 3 of the CoCoSo method determines the two strategies to aggregate weights of the criteria in the decision process. The table below presents this result.

**Table 4** – The aggregation strategies for CoCoSo.

| | L2 Students | | | | | | | L2 Students | | | | | |
|---|---|---|---|---|---|---|---|---|---|---|---|---|---|
| **Weight** | 0,45 | 0,18 | 0,1 | 0,1 | 0,18 | Si | **Weight** | 0,45 | 0,18 | 0,1 | 0,1 | 0,18 | Pi |
| **Student** | PC | DR | DC | PW | OP | | **Student** | PC | DR | DC | PW | OP | |
| **L201** | 0 | 0,0866 | 0,05 | 0 | 0 | **0,137** | L201 | 0 | 0,877 | 0,933 | 0 | 0 | **1,81** |
| **L202** | 0 | 0,0003 | 0,075 | 0 | 0,18 | **0,255** | L202 | 0 | 0,309 | 0,972 | 0 | 1 | **2,281** |
| **L203** | 0 | 0 | 0,075 | 0 | 0 | **0,075** | L203 | 0 | 0 | 0,972 | 0 | 0 | **0,972** |
| **L204** | 0 | 0 | 0,05 | 0 | 0,18 | **0,23** | L204 | 0 | 0 | 0,933 | 0 | 1 | **1,933** |
| **L205** | 0 | 0 | 0,025 | 0 | 0,18 | **0,205** | L205 | 0 | 0 | 0,871 | 0 | 1 | **1,871** |
| **L206** | 0 | 0 | 0,025 | 0 | 0,18 | **0,205** | L206 | 0 | 0 | 0,871 | 0 | 1 | **1,871** |
| **L207** | 0 | 0,1085 | 0,05 | 0 | 0,18 | **0,339** | L207 | 0 | 0,913 | 0,933 | 0 | 1 | **2,846** |
| **L208** | 0 | 0,1762 | 0,025 | 0,1 | 0 | **0,301** | L208 | 0 | 0,996 | 0,871 | 1 | 0 | **2,867** |
| **L209** | 0 | 0,18 | 0,05 | 0 | 0 | **0,23** | L209 | 0 | 1 | 0,933 | 0 | 0 | **1,933** |
| **L210** | 0,45 | 0 | 0,05 | 0 | 0 | **0,5** | L210 | 1 | 0 | 0,933 | 0 | 0 | **1,933** |
| **L211** | 0 | 0,0866 | 0,05 | 0 | 0 | **0,137** | L211 | 0 | 0,877 | 0,933 | 0 | 0 | **1,81** |
| **L212** | 0 | 0,0394 | 0,025 | 0 | 0,18 | **0,244** | L212 | 0 | 0,761 | 0,871 | 0 | 1 | **2,631** |
| **L213** | 0 | 0 | 0,1 | 0 | 0,18 | **0,28** | L213 | 0 | 0 | 1 | 0 | 1 | **2** |
| **L214** | 0 | 0,0003 | 0,05 | 0 | 0 | **0,05** | L214 | 0 | 0,309 | 0,933 | 0 | 0 | **1,242** |
| **L215** | 0 | 0,1549 | 0,05 | 0 | 0 | **0,205** | L215 | 0 | 0,973 | 0,933 | 0 | 0 | **1,906** |
| **L216** | 0 | 0,033 | 0,05 | 0 | 0 | **0,083** | L216 | 0 | 0,737 | 0,933 | 0 | 0 | **1,67** |
| **L217** | 0 | 0,033 | 0,025 | 0 | 0,18 | **0,238** | L217 | 0 | 0,737 | 0,871 | 0 | 1 | **2,607** |
| **L218** | 0,45 | 0 | 0,025 | 0 | 0,18 | **0,655** | L218 | 1 | 0 | 0,871 | 0 | 1 | **2,871** |
| **L219** | 0 | 0,0003 | 0,05 | 0,1 | 0,18 | **0,33** | L219 | 0 | 0,309 | 0,933 | 1 | 1 | **3,242** |
| **L220** | 0 | 0 | 0,025 | 0 | 0 | **0,025** | L220 | 0 | 0 | 0,871 | 0 | 0 | **0,871** |
| **L221** | 0 | 0 | 0 | 0 | 0 | **0** | L221 | 0 | 0 | 0 | 0 | 0 | **0** |
| **L222** | 0 | 0 | 0,05 | 0 | 0 | **0,05** | L222 | 0 | 0 | 0,933 | 0 | 0 | **0,933** |
| **L223** | 0 | 0,1549 | 0,075 | 0 | 0 | **0,23** | L223 | 0 | 0,973 | 0,972 | 0 | 0 | **1,945** |
| **L224** | 0 | 0,0671 | 0,05 | 0 | 0 | **0,117** | L224 | 0 | 0,837 | 0,933 | 0 | 0 | **1,77** |
| **L225** | 0 | 0 | 0,025 | 0 | 0 | **0,025** | L225 | 0 | 0 | 0,871 | 0 | 0 | **0,871** |
| **L226** | 0 | 0,1081 | 0,05 | 0 | 0 | **0,158** | L226 | 0 | 0,912 | 0,933 | 0 | 0 | **1,845** |

We continue the process by moving on to the fourth step of the CoCoSo method which consists of integrating the values of ***S*** and ***P***. So, we will have the result in the table below (**Table 5**). Integrating the value of **S** and **P** obtains the **Kia, Kib, Kic** and finally the **Ki** is determined to show the ranking of the students.

**Table 5** – The k values for CoCoSo

| λ       |       | 0,5     |       |         |
|---------|-------|---------|-------|---------|
| **Student** | **Kia** | **Kib** | **Kic** | **Ki** |
| L201    | 0,036 | #DIV/0! | 4,23  | #DIV/0! |
| L202    | 0,047 | #DIV/0! | 5,632 | #DIV/0! |
| L203    | 0,019 | #DIV/0! | 1,479 | #DIV/0! |
| L204    | 0,04  | #DIV/0! | 4,491 | #DIV/0! |
| L205    | 0,039 | #DIV/0! | 4,321 | #DIV/0! |
| L206    | 0,039 | #DIV/0! | 4,321 | #DIV/0! |
| L207    | 0,059 | #DIV/0! | 7,417 | #DIV/0! |
| L208    | 0,059 | #DIV/0! | 7,551 | #DIV/0! |
| L209    | 0,04  | #DIV/0! | 4,491 | #DIV/0! |
| L210    | 0,045 | #DIV/0! | 4,031 | #DIV/0! |
| L211    | 0,036 | #DIV/0! | 4,23  | #DIV/0! |
| L212    | 0,053 | #DIV/0! | 6,846 | #DIV/0! |
| L213    | 0,042 | #DIV/0! | 4,634 | #DIV/0! |
| L214    | 0,024 | #DIV/0! | 2,443 | #DIV/0! |
| L215    | 0,039 | #DIV/0! | 4,443 | #DIV/0! |
| L216    | 0,033 | #DIV/0! | 3,844 | #DIV/0! |
| L217    | 0,053 | #DIV/0! | 6,775 | #DIV/0! |
| L218    | 0,065 | #DIV/0! | 6,961 | #DIV/0! |
| L219    | 0,066 | #DIV/0! | 8,78  | #DIV/0! |
| L220    | 0,017 | #DIV/0! | 1,22  | #DIV/0! |
| L221    | 0     | #DIV/0! | -1,7  | #DIV/0! |
| L222    | 0,018 | #DIV/0! | 1,39  | #DIV/0! |
| L223    | 0,04  | #DIV/0! | 4,532 | #DIV/0! |
| L224    | 0,035 | #DIV/0! | 4,129 | #DIV/0! |
| L225    | 0,017 | #DIV/0! | 1,22  | #DIV/0! |
| L226    | 0,037 | #DIV/0! | 4,315 | #DIV/0! |

At this stage we also encountered a second error of the CoCoSo method algorithm. It will be impossible to obtain the values of $K_{ib}$ and obviously the rest of the process. The origin of this error comes from the presence of student **L221** in the decision matrix because it has *the minimum value on all the criteria*. Thus, we can infer that the second error on the CoCoSo method algorithm occurs when an alternative has minimum values on all criteria. While in the same case, it will always be possible to apply other multi-criteria decision-making methods such as AHP, WSM, PROMETHEE Rasoanaivo et Zaraté, (2023) and obtain the ranking score.

*3.3. CoCoFISo algorithm*

Being aware of the above mentioned errors in the CoCoSo method, we have carried out an analysis to resolve those errors. The objective of the CoCoFISo method is finding an ideal compromised solution to improve the algorithm of the CoCoSo method. To do that, we will modify the two components of the CoCoSo algorithm which present errors and will explain them separately.

### 3.3.1. Modification on normalization part

The extracted error on the normalization matrix is caused by the difference between $\max_i x_{ij}$ and $\min_i x_{ij}$ in the denominator of formulas (2) and (3). This happens when a criterion has the same value on all the alternatives because the denominator becomes zero. So, mathematically speaking, dividing a number by zero is impossible. CoCoSo is a method based on a combination of compromise attitudes Yazdani et al., (2019). In this case, to find an ideal solution, we propose to use the normalization method of the TOPSIS method Hwang et Yoon, (1981) to avoid the problem at the level of the denominator of the CoCoSo method. We have already used the TOPSIS method and found no error on the normalization of the matrix Rasoanaivo, (2023). This calculation method is capable of processing different real cases of multi-criteria decisions. Thus, for the CoCoFISo method we substitute the normalization process by allowing the following formula:

$$r_{ij} = \frac{x_{ij}}{\sqrt{\sum_{i=1}^{m}(x_{ij})^2}} \quad (10)$$

### 3.3.2. Modifying the $K_{ib}$: evaluation strategy

While using the CoCoSo method, we observed that the strategy aggregation $K_{ib}$ caused an error following the values of $S_i$ and $P_i$ which become zero if the alternative takes minimum values on all the criteria. Thus, the algorithm is not able to calculate the $K_{ib}$ because the denominator of this formula seeks for the minimum value among $S_i$ and $P_i$ and obviously, this minimum will fall to zero. And as before we mentioned, the algorithm will be stopped. To resolve this, CoCoFISo method operates the $K_{ib}$ using below formula to resolve this situation:

$$k_{ib} = \left( \frac{S_i + P_i}{1 + \frac{S_i}{1+S_i} + \frac{P_i}{1+P_i}} \right) \quad (11)$$

In the next section, we apply the CoCoFISo method for the student's residence allocation problem.

### 3.3.3. Comparative flow chart of formulas between CoCoSo and CoCoFISo

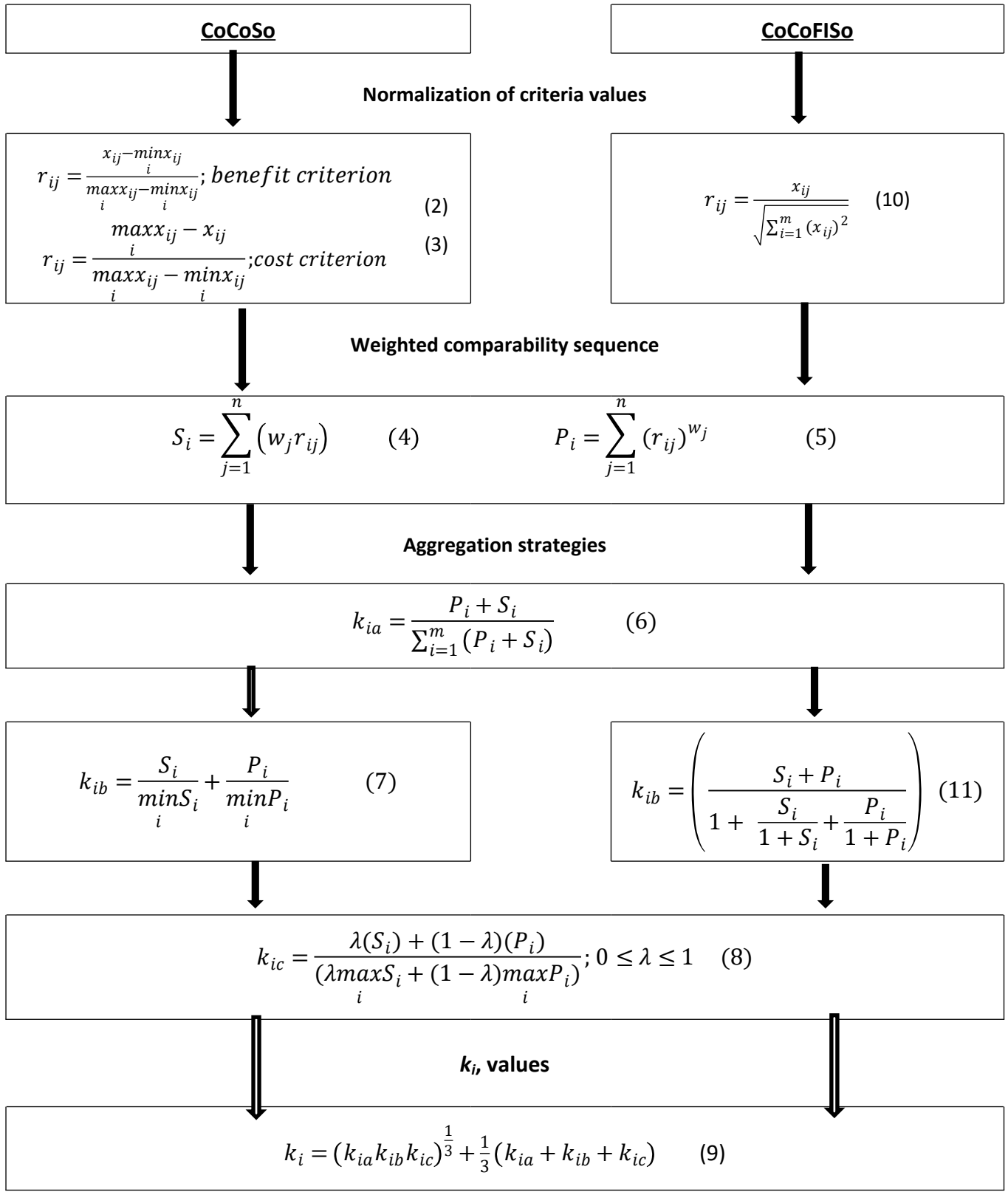

## 4. Application of the CoCoFISo method

Let's return to the example of allocating students in L1 and L2 by applying the CoCoFISo algorithm step by step. We will apply the two new formulas integrated into the classic CoCoSo.

### *4.1. Normalized matrix*

Here we consider the new normalization tool based on the TOPSIS method. We will have the result in the table below.

**Table 6** – The new normalized matrix for CoCoFISo

| L1 Students | | | | | | L2 Students | | | | | |
|---|---|---|---|---|---|---|---|---|---|---|---|
| Student | PC | DR | DC | PW | OP | Student | PC | DR | DC | PW | OP |
| L101 | 0,17 | 0,07 | 0,17 | 0,19 | 0,11 | L201 | 0,18 | 0,23 | 0,2 | 0,18 | 0,13 |
| L102 | 0,17 | 0,07 | 0,28 | 0,19 | 0,22 | L202 | 0,18 | 0,03 | 0,27 | 0,18 | 0,27 |
| L103 | 0,17 | 0,07 | 0,33 | 0,19 | 0,11 | L203 | 0,18 | 0,03 | 0,27 | 0,18 | 0,13 |
| L104 | 0,17 | 0,07 | 0,11 | 0,19 | 0,22 | L204 | 0,18 | 0,03 | 0,2 | 0,18 | 0,27 |
| L105 | 0,17 | 0,07 | 0,17 | 0,19 | 0,11 | L205 | 0,18 | 0,03 | 0,13 | 0,18 | 0,27 |
| L106 | 0,33 | 0,07 | 0,28 | 0,19 | 0,22 | L206 | 0,18 | 0,03 | 0,13 | 0,18 | 0,27 |
| L107 | 0,17 | 0,07 | 0,22 | 0,19 | 0,22 | L207 | 0,18 | 0,28 | 0,2 | 0,18 | 0,27 |
| L108 | 0,17 | 0,07 | 0,11 | 0,19 | 0,11 | L208 | 0,18 | 0,43 | 0,13 | 0,35 | 0,13 |
| L109 | 0,17 | 0,07 | 0,17 | 0,19 | 0,11 | L209 | 0,18 | 0,44 | 0,2 | 0,18 | 0,13 |
| L110 | 0,17 | 0,6 | 0,06 | 0,19 | 0,32 | L210 | 0,35 | 0,03 | 0,2 | 0,18 | 0,13 |
| L111 | 0,17 | 0,07 | 0,17 | 0,19 | 0,11 | L211 | 0,18 | 0,23 | 0,2 | 0,18 | 0,13 |
| L112 | 0,33 | 0,07 | 0,22 | 0,19 | 0,22 | L212 | 0,18 | 0,12 | 0,13 | 0,18 | 0,27 |
| L113 | 0,17 | 0,07 | 0,11 | 0,19 | 0,11 | L213 | 0,18 | 0,03 | 0,33 | 0,18 | 0,27 |
| L114 | 0,17 | 0,07 | 0,22 | 0,19 | 0,11 | L214 | 0,18 | 0,03 | 0,2 | 0,18 | 0,13 |
| L115 | 0,17 | 0,07 | 0,06 | 0,19 | 0,22 | L215 | 0,18 | 0,38 | 0,2 | 0,18 | 0,13 |
| L116 | 0,17 | 0,23 | 0,11 | 0,19 | 0,32 | L216 | 0,18 | 0,1 | 0,2 | 0,18 | 0,13 |
| L117 | 0,17 | 0,07 | 0,33 | 0,19 | 0,11 | L217 | 0,18 | 0,1 | 0,13 | 0,18 | 0,27 |
| L118 | 0,17 | 0,07 | 0,28 | 0,19 | 0,11 | L218 | 0,35 | 0,03 | 0,13 | 0,18 | 0,27 |
| L119 | 0,33 | 0,07 | 0,11 | 0,19 | 0,11 | L219 | 0,18 | 0,03 | 0,2 | 0,35 | 0,27 |
| L120 | 0,17 | 0,07 | 0,06 | 0,19 | 0,11 | L220 | 0,18 | 0,03 | 0,13 | 0,18 | 0,13 |
| L121 | 0,17 | 0,07 | 0,22 | 0,19 | 0,22 | L221 | 0,18 | 0,03 | 0,07 | 0,18 | 0,13 |
| L122 | 0,17 | 0,07 | 0,28 | 0,19 | 0,22 | L222 | 0,18 | 0,03 | 0,2 | 0,18 | 0,13 |
| L123 | 0,17 | 0,26 | 0,06 | 0,19 | 0,11 | L223 | 0,18 | 0,38 | 0,27 | 0,18 | 0,13 |
| L124 | 0,17 | 0,07 | 0,17 | 0,19 | 0,32 | L224 | 0,18 | 0,18 | 0,2 | 0,18 | 0,13 |
| L125 | 0,17 | 0,6 | 0,22 | 0,19 | 0,22 | L225 | 0,18 | 0,03 | 0,13 | 0,18 | 0,13 |
| L126 | 0,17 | 0,26 | 0,11 | 0,19 | 0,32 | L226 | 0,18 | 0,27 | 0,2 | 0,18 | 0,13 |
| L127 | 0,17 | 0,07 | 0,11 | 0,19 | 0,11 | | | | | | |

By applying the new method of calculating the normalized matrix, we note that the error on the **PW** criterion of L1 students has been corrected. The application of this algorithm also made it possible to easily calculate the normalized matrix of L2 students. They certify that the real cases were treated by correcting this algorithm.

### 4.2. Determination of the two strategies to aggregate weights

Application of formula 11 to find the aggregation strategies in CoCoFISo is shown in **Table 7** where the Pi and Si values for both groups of students are detected.

**Table 7** – The aggregation strategies for CoCoFISo method

| L1 students | | | L2 students | | |
|---|---|---|---|---|---|
| Student | Si | Pi | Student | Si | Pi |
| L101 | 0,142 | 3,411 | L201 | 0,182 | 3,612 |
| L102 | 0,172 | 3,544 | L202 | 0,178 | 3,498 |
| L103 | 0,158 | 3,471 | L203 | 0,153 | 3,404 |
| L104 | 0,156 | 3,469 | L204 | 0,171 | 3,471 |
| L105 | 0,142 | 3,411 | L205 | 0,164 | 3,437 |
| L106 | 0,247 | 3,707 | L206 | 0,164 | 3,437 |
| L107 | 0,167 | 3,524 | L207 | 0,215 | 3,733 |
| L108 | 0,136 | 3,378 | L208 | 0,230 | 3,732 |
| L109 | 0,142 | 3,411 | L209 | 0,220 | 3,709 |
| L110 | 0,266 | 3,771 | L210 | 0,226 | 3,547 |
| L111 | 0,142 | 3,411 | L211 | 0,182 | 3,612 |
| L112 | 0,242 | 3,688 | L212 | 0,18 | 3,588 |
| L113 | 0,136 | 3,378 | L213 | 0,184 | 3,516 |
| L114 | 0,147 | 3,435 | L214 | 0,147 | 3,381 |
| L115 | 0,150 | 3,413 | L215 | 0,210 | 3,688 |
| L116 | 0,204 | 3,679 | L216 | 0,160 | 3,513 |
| L117 | 0,158 | 3,471 | L217 | 0,178 | 3,572 |
| L118 | 0,153 | 3,455 | L218 | 0,244 | 3,605 |
| L119 | 0,211 | 3,541 | L219 | 0,189 | 3,533 |
| L120 | 0,131 | 3,326 | L220 | 0,140 | 3,345 |
| L121 | 0,167 | 3,524 | **L221** | **0,133** | **3,290** |
| L122 | 0,172 | 3,544 | L222 | 0,147 | 3,379 |
| L123 | 0,166 | 3,497 | L223 | 0,217 | 3,712 |
| L124 | 0,181 | 3,557 | L224 | 0,174 | 3,583 |
| L125 | 0,263 | 3,825 | L225 | 0,140 | 3,345 |
| L126 | 0,210 | 3,697 | L226 | 0,191 | 3,640 |
| L127 | 0,136 | 3,378 | | | |

We noticed that no value of **Si** and **Pi** falls on zero nor of students in L1 and L2. On classic CoCoSo, the value of S and P of the L2 student referenced L221 was zero. On the other hand, with application of CoCoFISo these values have been corrected. This is a positive feedback and signal for the rest of our decision-making process.

The new aggregation strategy by applying the new $K_{ib}$ formula will result in accurate values which are presented in **Table 8**.

**Table 8** – k values for new improved CoCoFISo method

| L1 students | | | | L2 students | | | |
|---|---|---|---|---|---|---|---|
| **Student** | **Kia** | **kib** | **kic** | **Student** | **Kia** | **kib** | **kic** |
| L101 | 0,036 | 0,438 | 0,868 | L201 | 0,039 | 0,442 | 0,954 |
| L102 | 0,037 | 0,441 | 0,908 | L202 | 0,038 | 0,440 | 0,924 |
| L103 | 0,036 | 0,439 | 0,887 | L203 | 0,037 | 0,438 | 0,894 |
| L104 | 0,036 | 0,439 | 0,886 | L204 | 0,038 | 0,440 | 0,915 |
| L105 | 0,036 | 0,438 | 0,868 | L205 | 0,037 | 0,439 | 0,905 |
| L106 | 0,040 | 0,444 | 0,966 | L206 | 0,037 | 0,439 | 0,905 |
| L107 | 0,037 | 0,440 | 0,902 | L207 | 0,041 | 0,444 | 0,992 |
| L108 | 0,035 | 0,438 | 0,859 | L208 | 0,041 | 0,444 | 0,996 |
| L109 | 0,036 | 0,438 | 0,868 | L209 | 0,041 | 0,444 | 0,988 |
| L110 | 0,041 | 0,445 | 0,986 | L210 | 0,039 | 0,441 | 0,948 |
| L111 | 0,036 | 0,438 | 0,868 | L211 | 0,039 | 0,442 | 0,954 |
| L112 | 0,039 | 0,444 | 0,960 | L212 | 0,039 | 0,441 | 0,947 |
| L113 | 0,035 | 0,438 | 0,859 | L213 | 0,038 | 0,440 | 0,930 |
| L114 | 0,036 | 0,439 | 0,875 | L214 | 0,037 | 0,438 | 0,887 |
| L115 | 0,036 | 0,438 | 0,871 | L215 | 0,040 | 0,443 | 0,980 |
| L116 | 0,039 | 0,443 | 0,949 | L216 | 0,038 | 0,440 | 0,923 |
| L117 | 0,036 | 0,439 | 0,887 | L217 | 0,039 | 0,441 | 0,942 |
| L118 | 0,036 | 0,439 | 0,881 | L218 | 0,040 | 0,443 | 0,967 |
| L119 | 0,038 | 0,441 | 0,917 | L219 | 0,039 | 0,441 | 0,936 |
| L120 | 0,035 | 0,437 | 0,845 | L220 | 0,036 | 0,437 | 0,876 |
| L121 | 0,037 | 0,440 | 0,902 | L221 | 0,035 | 0,436 | 0,861 |
| L122 | 0,037 | 0,441 | 0,908 | L222 | 0,036 | 0,438 | 0,886 |
| L123 | 0,037 | 0,440 | 0,895 | L223 | 0,041 | 0,444 | 0,988 |
| L124 | 0,038 | 0,441 | 0,913 | L224 | 0,039 | 0,441 | 0,944 |
| L125 | 0,041 | 0,446 | 0,999 | L225 | 0,036 | 0,437 | 0,876 |
| L126 | 0,039 | 0,443 | 0,955 | L226 | 0,040 | 0,442 | 0,963 |
| L127 | 0,035 | 0,438 | 0,859 | | | | |

Following the change in $K_{ib}$'s algorithm, we notice that its values for L2 students are computable. We no longer have a problem in any case. The final rank will be obtained from the $K_i$ value. We will find these values and the ranking of the students in the table below.

**Table 9** – the ranking of the L1 and L2 groups

| L1 Students | | | L2 Students | | |
| --- | --- | --- | --- | --- | --- |
| Student | ki | Ranking | Student | ki | Ranking |
| L125 | 0,758 | 1 | L208 | 0,756 | 1 |
| L110 | 0,751 | 2 | L207 | 0,754 | 2 |
| L106 | 0,740 | 3 | L223 | 0,75 | 3 |
| L112 | 0,737 | 4 | L209 | 0,752 | 4 |
| L126 | 0,734 | 5 | L215 | 0,747 | 5 |
| L116 | 0,731 | 6 | L218 | 0,740 | 6 |
| L119 | 0,713 | 7 | L226 | 0,738 | 7 |
| L124 | 0,711 | 8 | L201 | 0,733 | 8 |
| L102 | 0,708 | 9 | L211 | 0,733 | 8 |
| L122 | 0,708 | 9 | L210 | 0,730 | 10 |
| L107 | 0,705 | 11 | L212 | 0,729 | 11 |
| L121 | 0,705 | 11 | L224 | 0,728 | 12 |
| L123 | 0,701 | 13 | L217 | 0,727 | 13 |
| L103 | 0,696 | 14 | L219 | 0,723 | 14 |
| L117 | 0,696 | 14 | L213 | 0,720 | 15 |
| L104 | 0,695 | 16 | L202 | 0,716 | 16 |
| L118 | 0,693 | 17 | L216 | 0,716 | 17 |
| L114 | 0,690 | 18 | L204 | 0,712 | 18 |
| 115 | 0,687 | 19 | L205 | 0,706 | 19 |
| L101 | 0,686 | 20 | L206 | 0,706 | 19 |
| L105 | 0,686 | 20 | L203 | 0,700 | 21 |
| L109 | 0,686 | 20 | L214 | 0,695 | 22 |
| L111 | 0,686 | 20 | L222 | 0,695 | 23 |
| L108 | 0,680 | 24 | L220 | 0,689 | 24 |
| L113 | 0,680 | 24 | L225 | 0,689 | 24 |
| L127 | 0,680 | 24 | L221 | 0,681 | 26 |
| L120 | 0,672 | 27 | | | |

We reached the final stage and obtained the rank of the students of each level of study. In MCDM studies, usually several tests will be handled to validate and check the stability of the results.

## 5. Testing CoCoFISo

To be able to test the ranks obtained by the CoCoFISo method, we carried out two analyses. First, a ranking comparison with other multi-criteria decision-making methods was performed. Following this comparison, the rank correlation coefficients were calculated to be able to validate the ranks obtained by the CoCoFISo method. Secondly, a study of modifying the weights of the criteria (sensitivity analysis) was implemented to guarantee the weight management policy of the CoCoFISo method. We present below these two analyses for testing ranks for the CoCoFISo method.

### *5.1. Comparing ranks with other methods*

To compare the ranks obtained by the CoCoFISo method, we proceeded to the application of the PROMETHEE, WSM and TOPSIS methods based on the same decision matrices of the students in L1 and L2. Thus, **Figure 1** presents this ranking comparison for L1 students and L2 students.

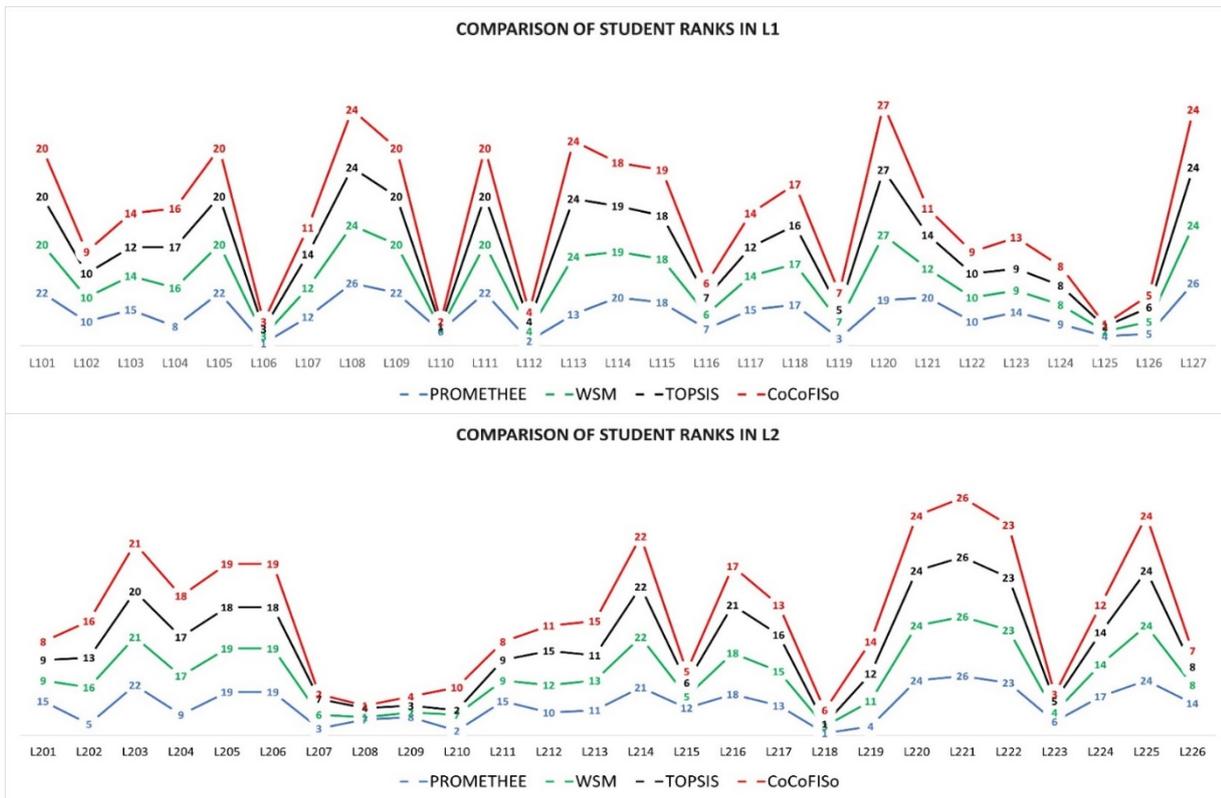

**Figure 1 –** Ranking comparison of CoCoFISo

To go further on this analysis, we have presented another **Figure 2** which addresses the equality and difference statistics of the ranks of the CoCoFISo method compared to the other methods. We can observe that the percentage of equality of ranks is generally low compared to the percentage of difference of ranks on all methods except that of the WSM method for L1 students.

This presentation makes it easier for us to read the comparison of ranks obtained in the previous figure. Rank equality means that the CoCoFISo method was able to keep the rank of a student resulting from another method. The difference in rank tells us that the CoCoFISo method was unable to keep the rank of a student resulting from another method. It is normal if the ranks obtained are different because each method has its own specificity.

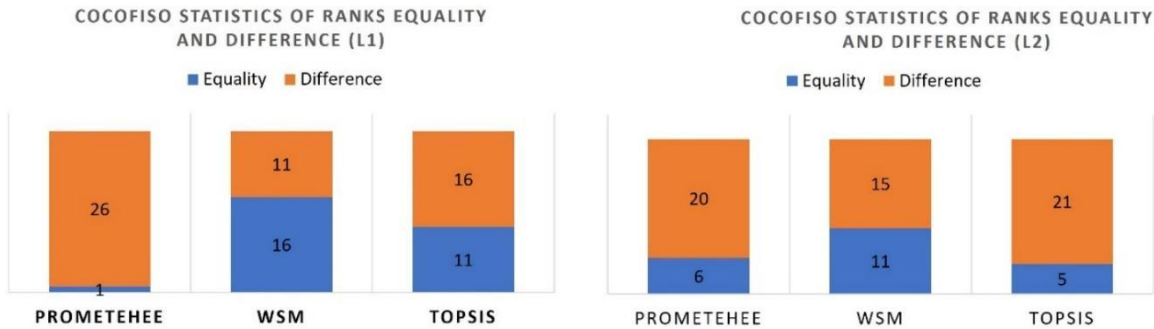

**Figure 2** – New CoCoSo method statistics

Now, by the obtained rankings of these different methods, the following table presents the correlation coefficients of the Spearman and Kendall ranks of the CoCoFISo method compared to the PROMETHEE, WSM and TOPSIS methods.

The correlation coefficient is a unitless value between -1 and 1. The closer is to zero, the weaker the linear relationship. Positive values indicate a positive correlation when the values of both variables tend to increase together. Negative values indicate a negative correlation when the values of one variable tend to increase and the values of the other variable decrease. The values 1 and -1 each represent "perfect" correlations, positive and negative respectively.

**Table 10** – the comparative Correlation coefficient for L1 and L2 based on various rankings

| L1 Students | PROMETHEE | WSM | TOPSIS | |
|---|---|---|---|---|
| CoCoFISo | 0,87 | 0,99 | 0,98 | **Spearman** |
| | 0,75 | 0,96 | 0,92 | **Kendall** |
| **L2 Students** | **PROMETHEE** | **WSM** | **TOPSIS** | |
| CoCoFISo | 0,76 | 0,98 | 0,93 | **Spearman** |
| | 0,63 | 0,73 | 0,80 | **Kendall** |

It is seen that the coefficients range from 63% to 99%. In terms of correlation, once the value is greater than 70%, we can state a good agreement for the correlation. Also, as the values of the correlation coefficients obtained generally tend towards 1, these mean that the ranks obtained by the CoCoFISo method have the same meaning as those of the PROMETHEE, WSM and TOPSIS methods. Technically speaking, the result of the CoCoFISo method is validated.

*5.2. Changing criteria weights*

One of the specific features of multi-criteria decision-making methods is the possibility of management based on the weights of the criteria, which is called weight sensitivity analysis. To take advantage of this concept, we carried out a second analysis of the CoCoFISo method by applying 20 different categories of weights for the criteria presented in the following table.

**Table 11** – The sensitivity analysis for weight replacement

| Wi | PC | DR | DC | PW | OP |
|---|---|---|---|---|---|
| W1 | 0,45 | 0,1 | 0,18 | 0,18 | 0,1 |
| W2 | 0,45 | 0,18 | 0,1 | 0,18 | 0,1 |
| W3 | 0,45 | 0,1 | 0,1 | 0,18 | 0,18 |
| W4 | 0,45 | 0,18 | 0,18 | 0,1 | 0,1 |
| W5 | 0,18 | 0,45 | 0,1 | 0,1 | 0,18 |
| W6 | 0,18 | 0,45 | 0,18 | 0,1 | 0,1 |
| W7 | 0,1 | 0,45 | 0,18 | 0,18 | 0,1 |
| W8 | 0,1 | 0,45 | 0,1 | 0,18 | 0,18 |
| W9 | 0,18 | 0,1 | 0,45 | 0,1 | 0,18 |
| W10 | 0,1 | 0,18 | 0,45 | 0,1 | 0,18 |
| W11 | 0,1 | 0,1 | 0,45 | 0,18 | 0,18 |
| W12 | 0,18 | 0,18 | 0,45 | 0,1 | 0,1 |
| W13 | 0,18 | 0,18 | 0,1 | 0,45 | 0,1 |
| W14 | 0,1 | 0,18 | 0,18 | 0,45 | 0,1 |
| W15 | 0,1 | 0,1 | 0,18 | 0,45 | 0,18 |
| W16 | 0,18 | 0,1 | 0,1 | 0,45 | 0,18 |
| W17 | 0,18 | 0,1 | 0,1 | 0,18 | 0,45 |
| W18 | 0,1 | 0,1 | 0,18 | 0,18 | 0,45 |
| W19 | 0,18 | 0,18 | 0,1 | 0,1 | 0,45 |
| W20 | 0,1 | 0,18 | 0,1 | 0,18 | 0,45 |

The objective of this analysis is to change the priority of the criteria. We assigned a priority weight (0.45) four times to each criterion (PC, DR, DC, PW, OP) and modified at the same time the weights of the other criteria (0.18 or 0.10). Thus, the result of the rankings according to the application of the different weights (W1, W2, …, W20) for **L1 students** and **L2 students** are given in the **tables 12 and 13**. Based on the observed results and CoCoFISo method, it is validated that the rankings will be affected too slightly and almost there are no considerable changes in ranking outcomes by varying the criteria weights. For instance, L125 is introduced in 15 tests as the best selected student, and 2$^{nd}$ ranking in the rest of the tests for L1 students. The same condition is screened in the L2 students group where L208 has the majority (15 times scored as best or second item) of 1th ranking in all 20 tests.

*Expert Systems with Applications,* https://doi.org/10.1016/j.eswa.2024.124079**Table 12 -** Ranking according to the sensitivity analysis test for L1 students

| Criteria | PC | PC | PC | PC | DR | DR | DR | DR | DC | DC | DC | DC | PW | PW | PW | PW | OP | OP | OP | OP |
|---|---|---|---|---|---|---|---|---|---|---|---|---|---|---|---|---|---|---|---|---|
| Student | W1 | W2 | W3 | W4 | W5 | W6 | W7 | W8 | W9 | W10 | W11 | W12 | W13 | W14 | W15 | W16 | W17 | W18 | W19 | W20 |
| L101 | 18 | 19 | 20 | 19 | 20 | 19 | 19 | 20 | 18 | 17 | 17 | 17 | 19 | 19 | 18 | 20 | 20 | 19 | 20 | 20 |
| L102 | 8 | 8 | 9 | 8 | 9 | 8 | 8 | 9 | 4 | 5 | 3 | 7 | 7 | 7 | 7 | 8 | 8 | 8 | 8 | 8 |
| L103 | 10 | 14 | 13 | 10 | 14 | 10 | 10 | 13 | 6 | 9 | 6 | 5 | 14 | 9 | 12 | 13 | 15 | 12 | 16 | 15 |
| L104 | 17 | 17 | 15 | 18 | 16 | 18 | 18 | 15 | 17 | 16 | 16 | 22 | 17 | 18 | 15 | 15 | 12 | 14 | 12 | 12 |
| L105 | 18 | 19 | 20 | 19 | 20 | 19 | 19 | 20 | 18 | 17 | 17 | 17 | 19 | 19 | 18 | 20 | 20 | 19 | 20 | 20 |
| L106 | 1 | 3 | 1 | 2 | 5 | 5 | 5 | 6 | 2 | 3 | 2 | 2 | 5 | 5 | 3 | 3 | 5 | 5 | 5 | 6 |
| L107 | 12 | 12 | 11 | 12 | 11 | 12 | 12 | 11 | 11 | 11 | 11 | 12 | 11 | 11 | 10 | 10 | 10 | 10 | 10 | 10 |
| L108 | 23 | 24 | 24 | 23 | 24 | 23 | 23 | 24 | 23 | 23 | 23 | 23 | 24 | 23 | 24 | 24 | 24 | 24 | 24 | 24 |
| L109 | 18 | 19 | 20 | 19 | 20 | 19 | 19 | 20 | 18 | 17 | 17 | 17 | 19 | 19 | 18 | 20 | 20 | 19 | 20 | 20 |
| L110 | 4 | 2 | 4 | 3 | 2 | 2 | 2 | 2 | 9 | 2 | 9 | 3 | 2 | 2 | 2 | 2 | 1 | 2 | 1 | 1 |
| L111 | 18 | 19 | 20 | 19 | 20 | 19 | 19 | 20 | 18 | 17 | 17 | 17 | 19 | 19 | 18 | 20 | 20 | 19 | 20 | 20 |
| L112 | 3 | 4 | 3 | 4 | 6 | 6 | 6 | 7 | 3 | 7 | 5 | 4 | 6 | 6 | 6 | 5 | 6 | 7 | 6 | 7 |
| L113 | 23 | 24 | 24 | 23 | 24 | 23 | 23 | 24 | 23 | 23 | 23 | 23 | 24 | 23 | 24 | 24 | 24 | 24 | 24 | 24 |
| L114 | 16 | 18 | 18 | 17 | 18 | 17 | 16 | 18 | 15 | 15 | 15 | 15 | 18 | 16 | 16 | 18 | 19 | 16 | 19 | 19 |
| L115 | 25 | 23 | 19 | 25 | 19 | 25 | 25 | 19 | 25 | 25 | 25 | 25 | 23 | 25 | 23 | 19 | 14 | 17 | 15 | 14 |
| L116 | 7 | 6 | 6 | 6 | 4 | 4 | 4 | 4 | 10 | 8 | 10 | 10 | 4 | 4 | 5 | 6 | 4 | 4 | 4 | 4 |
| L117 | 10 | 14 | 13 | 10 | 14 | 10 | 10 | 13 | 6 | 9 | 6 | 5 | 14 | 9 | 12 | 13 | 15 | 12 | 16 | 15 |
| L118 | 15 | 16 | 16 | 15 | 17 | 15 | 15 | 16 | 14 | 14 | 14 | 11 | 16 | 14 | 14 | 16 | 17 | 15 | 18 | 17 |
| L119 | 6 | 7 | 7 | 7 | 13 | 16 | 17 | 17 | 16 | 22 | 21 | 16 | 13 | 17 | 17 | 12 | 13 | 18 | 14 | 18 |
| L120 | 26 | 26 | 26 | 26 | 26 | 26 | 26 | 26 | 26 | 26 | 26 | 26 | 26 | 26 | 26 | 26 | 26 | 26 | 26 | 26 |
| L121 | 12 | 12 | 11 | 12 | 11 | 12 | 12 | 11 | 11 | 11 | 11 | 12 | 11 | 11 | 10 | 10 | 10 | 10 | 10 | 10 |
| L122 | 8 | 8 | 9 | 8 | 9 | 8 | 8 | 9 | 4 | 5 | 3 | 7 | 7 | 7 | 7 | 8 | 8 | 8 | 8 | 8 |
| L123 | 22 | 11 | 17 | 16 | 7 | 7 | 7 | 5 | 22 | 21 | 22 | 21 | 10 | 15 | 22 | 17 | 18 | 23 | 13 | 13 |
| L124 | 14 | 10 | 8 | 14 | 8 | 14 | 14 | 8 | 13 | 13 | 13 | 14 | 9 | 13 | 9 | 7 | 7 | 6 | 7 | 5 |
| L125 | 2 | 1 | 2 | 1 | 1 | 1 | 1 | 1 | 1 | 1 | 1 | 1 | 1 | 1 | 1 | 1 | 2 | 1 | 2 | 2 |
| L126 | 5 | 5 | 5 | 5 | 3 | 3 | 3 | 3 | 8 | 4 | 8 | 9 | 3 | 3 | 4 | 4 | 3 | 3 | 3 | 3 |
| L127 | 23 | 24 | 24 | 23 | 24 | 23 | 23 | 24 | 23 | 23 | 23 | 23 | 24 | 23 | 24 | 24 | 24 | 24 | 24 | 24 |

Now let's move on to the result of changing the criteria weights for L2 students in the table below.

*Available online 20 April 2024, 124079*

Table 13 - Ranking according to the sensitivity analysis test for L2 students

| | L2 Students | | | | | | | | | | | | | | | | | | | |
|---|---|---|---|---|---|---|---|---|---|---|---|---|---|---|---|---|---|---|---|---|
| | PC | PC | PC | PC | DR | DR | DR | DR | DC | DC | DC | DC | PW | PW | PW | PW | OP | OP | OP | OP |
| Student | W1 | W2 | W3 | W4 | W5 | W6 | W7 | W8 | W9 | W10 | W11 | W12 | W13 | W14 | W15 | W16 | W17 | W18 | W19 | W20 |
| L201 | 10 | 8 | 11 | 7 | 7 | 7 | 7 | 7 | 10 | 8 | 10 | 7 | 8 | 8 | 9 | 10 | 14 | 14 | 11 | 10 |
| L202 | 14 | 17 | 16 | 17 | 16 | 17 | 15 | 16 | 7 | 11 | 8 | 11 | 17 | 15 | 11 | 15 | 10 | 6 | 14 | 14 |
| L203 | 19 | 19 | 21 | 19 | 21 | 19 | 19 | 21 | 19 | 17 | 17 | 17 | 19 | 19 | 19 | 21 | 21 | 21 | 21 | 21 |
| L204 | 18 | 18 | 17 | 18 | 18 | 18 | 18 | 17 | 16 | 16 | 14 | 19 | 18 | 18 | 16 | 17 | 12 | 12 | 16 | 16 |
| L205 | 20 | 20 | 19 | 20 | 19 | 20 | 20 | 19 | 20 | 20 | 20 | 22 | 20 | 20 | 20 | 19 | 16 | 17 | 17 | 17 |
| L206 | 20 | 20 | 19 | 20 | 19 | 20 | 20 | 19 | 20 | 20 | 20 | 22 | 20 | 20 | 20 | 19 | 16 | 17 | 17 | 17 |
| L207 | 3 | 4 | 3 | 4 | 5 | 5 | 5 | 5 | 2 | 2 | 2 | 3 | 4 | 4 | 3 | 3 | 1 | 1 | 1 | 1 |
| L208 | 1 | 1 | 1 | 2 | 1 | 1 | 1 | 1 | 6 | 5 | 4 | 5 | 1 | 1 | 1 | 1 | 2 | 3 | 2 | 2 |
| L209 | 4 | 3 | 5 | 3 | 2 | 2 | 2 | 2 | 4 | 3 | 5 | 2 | 3 | 3 | 5 | 5 | 9 | 7 | 4 | 4 |
| L210 | 6 | 10 | 6 | 10 | 17 | 16 | 17 | 18 | 14 | 19 | 19 | 14 | 16 | 17 | 18 | 16 | 19 | 20 | 20 | 20 |
| L211 | 10 | 8 | 11 | 7 | 7 | 7 | 7 | 7 | 10 | 8 | 10 | 7 | 8 | 8 | 9 | 10 | 14 | 14 | 11 | 10 |
| L212 | 15 | 13 | 10 | 12 | 10 | 10 | 10 | 10 | 15 | 13 | 13 | 15 | 11 | 11 | 12 | 9 | 5 | 8 | 6 | 7 |
| L213 | 12 | 16 | 14 | 13 | 15 | 13 | 14 | 14 | 3 | 7 | 3 | 9 | 15 | 12 | 7 | 13 | 8 | 4 | 13 | 12 |
| L214 | 22 | 22 | 22 | 22 | 22 | 22 | 22 | 22 | 22 | 22 | 22 | 20 | 22 | 22 | 22 | 22 | 22 | 22 | 22 | 22 |
| L215 | 7 | 5 | 8 | 5 | 4 | 4 | 4 | 4 | 5 | 4 | 7 | 4 | 5 | 5 | 6 | 6 | 11 | 10 | 5 | 5 |
| L216 | 17 | 15 | 18 | 16 | 14 | 12 | 13 | 13 | 18 | 15 | 16 | 13 | 14 | 14 | 17 | 18 | 20 | 19 | 19 | 19 |
| L217 | 16 | 14 | 13 | 14 | 11 | 11 | 12 | 11 | 17 | 14 | 15 | 16 | 12 | 13 | 14 | 12 | 7 | 9 | 7 | 8 |
| L218 | 5 | 7 | 2 | 9 | 12 | 15 | 16 | 15 | 13 | 18 | 18 | 18 | 13 | 16 | 15 | 7 | 4 | 11 | 9 | 13 |
| L219 | 8 | 12 | 7 | 15 | 13 | 14 | 11 | 12 | 9 | 12 | 6 | 12 | 6 | 6 | 2 | 2 | 3 | 2 | 10 | 6 |
| L220 | 24 | 24 | 24 | 24 | 24 | 24 | 24 | 24 | 24 | 24 | 24 | 24 | 24 | 24 | 24 | 24 | 24 | 24 | 24 | 24 |
| L221 | 26 | 26 | 26 | 26 | 26 | 26 | 26 | 26 | 26 | 26 | 26 | 26 | 26 | 26 | 26 | 26 | 26 | 26 | 26 | 26 |
| L222 | 23 | 23 | 23 | 23 | 23 | 23 | 23 | 23 | 23 | 23 | 23 | 21 | 23 | 23 | 23 | 23 | 23 | 23 | 23 | 23 |
| L223 | 2 | 2 | 4 | 1 | 3 | 3 | 3 | 3 | 1 | 1 | 1 | 1 | 2 | 2 | 4 | 4 | 6 | 5 | 3 | 3 |
| L224 | 13 | 11 | 15 | 11 | 9 | 9 | 9 | 9 | 12 | 10 | 12 | 10 | 10 | 10 | 13 | 14 | 18 | 16 | 15 | 15 |
| L225 | 24 | 24 | 24 | 24 | 24 | 24 | 24 | 24 | 24 | 24 | 24 | 24 | 24 | 24 | 24 | 24 | 24 | 24 | 24 | 24 |
| L226 | 9 | 6 | 9 | 6 | 6 | 6 | 6 | 6 | 8 | 6 | 9 | 6 | 7 | 7 | 8 | 8 | 13 | 13 | 8 | 9 |

**6. Discussion, conclusion, and future works**

*6.1. Discussion of the proposed method*

In this section we present a brief discussion on the extracted results and finally provide a conclusion. Upon the results of the changes in the weights of the criteria, we found that for each prioritized criterion (PC, DR, DC, PW, OP) four different situations could be observed (1, 2, 3, 4). **Situation 1 (blue)** presents the percentage of students who have different ranks on the priority of a criterion four times. **Situation 2 (orange)** denotes the percentage of students who have half equal ranking when prioritizing a criterion four times. **Situation 3 (grey)** shows the percentage of students who have 3 equal ranks out of four. Lastly, **Situation 4 (yellow)** indicates the percentage of students who have stable ranks (4/4) on a criterion that has been prioritized four times.

Following this result, a rank stability analysis by prioritized criterion was carried out. The figure below shows the percentage of rank stability for students in L1.

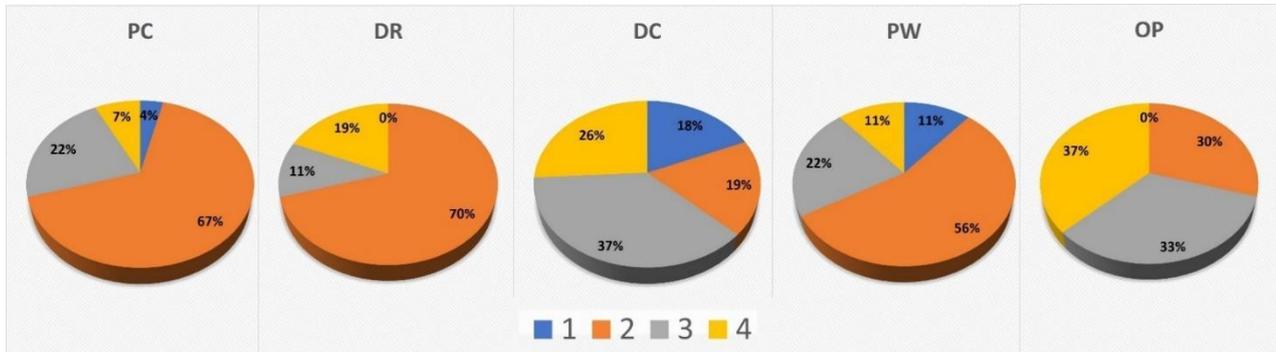

**Figure 3** – Rank stability analysis for CoCoFISo method (L1 students)

Compared to these situations we can draw that for the students in L1, the percentage of the differences of the ranks is low (0% to 18%) for each prioritization of the criteria. On the other hand, the percentage of equality of ranks is significant for each prioritized criterion. We can cite among other things that situation 2 has the highest percentage (30% to 67%), while situation 3 presents an average percentage (11% to 37%) and situation 4 leaves a percentage almost similar to situation 3 (7% to 37%). Therefore, this can guarantee that by changing the priority criteria for L1 students, the ranks are generally stable. Similarly, the four situations (1, 2, 3, 4) are also observed for L2 students which are presented in the figure below.

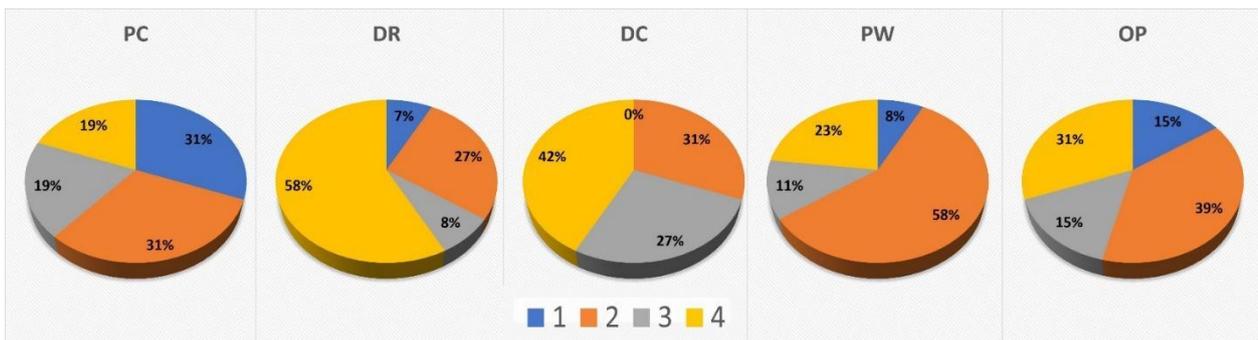

**Figure 4** - Rank stability analysis for CoCoFISo method (L2 students)

Thus, the percentage of students who have different ranks on the four prioritization of a criterion ranges from 7% to 31% for all criteria. Students who have half equal ranks are in the range of 27% to 58%. Students whose ranks are three-quarters (3/4) equal have a percentage of 8% to 27%. Finally, students who have equal ranks on all four modified weights range from 19% to 58%.

This analysis addresses us that by changing the criteria weights for L2 students, the ranks remained stable.

### *6.2. Conclusion*

In this study we have proposed a new version of the CoCoSo method according to some errors observed in the algorithm in special cases which we called CoCoFISo. Along with the extended version, the two real cases of students' residence allocation with two groups of students (L1 and L2) were assumed to show us the limit of the CoCoSo method. Compared to the results of the PROMETHEE, WSM and TOPSIS methods, the ranking obtained by the CoCoFISo method was validated according to the Spearman and Kendall correlation coefficients. Moreover, we have designed several tests by different criteria weights, and we have checked that the result of the CoCoFISo method was globally stable. The main advantage of CoCoFiSo is the correction of the first version of the algorithm. This new version is now usable without any exceptions and can solve any multi-criteria decision-making problem. In this article, the limit of the CoCoFISo method is that it was applied just for a case which is the allocation of student accommodation.

### *6.3. Future research work*

However, for future research, its application in various fields of knowledge will be interesting to know its strength. It is not always possible to obtain all data for such a method. Sometimes data is unavailable, sometimes it is uncertain. Researchers may in future develop the fuzzy CoCoFISo when uncertainty exists to implement new versions of fuzzy CoCoSo. This would allow us to consider uncertainty about data and scenario.